\documentclass[conference]{IEEEtran}
\IEEEoverridecommandlockouts
\usepackage{cite}
\usepackage{amsmath,amssymb,amsfonts}
\usepackage{algorithm}
\usepackage[noend]{algpseudocode}
\algnewcommand{\LeftComment}[1]{\Statex \hspace{4 mm} \(\triangleright\) #1}
\usepackage{graphicx}
\usepackage{textcomp}
\usepackage{xcolor}
\usepackage{rotating}
\def\BibTeX{{\rm B\kern-.05em{\sc i\kern-.025em b}\kern-.08em
    T\kern-.1667em\lower.7ex\hbox{E}\kern-.125emX}}
    
\definecolor{filler-michael}{rgb}{0.82, 0.82, 0.9}
\definecolor{filler-majiga}{rgb}{0.9, 0.82, 0.82}

\definecolor{filler-text}{rgb}{0.3, 0.3, 0.3}

\definecolor{unsurecolor}{rgb}{0.5, 0.5, 0.5}
\definecolor{lightgray}{rgb}{0.8, 0.8, 0.8}

\begin{document}

\title{ICDM 2019 Knowledge Graph Contest: Team UWA}

\author{
\IEEEauthorblockN{1\textsuperscript{st} Michael Stewart}
\IEEEauthorblockA{
\textit{The University of Western Australia}\\
Perth, Australia \\
michael.stewart@research.uwa.edu.au}
\and
\IEEEauthorblockN{2\textsuperscript{nd} Majigsuren Enkhsaikhan}
\IEEEauthorblockA{
\textit{The University of Western Australia}\\
Perth, Australia \\
majigsuren.enkhsaikhan@research.uwa.edu.au}
\and
\IEEEauthorblockN{3\textsuperscript{rd} Wei Liu}
\IEEEauthorblockA{
\textit{The University of Western Australia}\\
Perth, Australia \\
wei.liu@uwa.edu.au}
}
\maketitle



\section{Model}

\subsection{Introduction}

We begin our report by discussing the challenges we experienced and the motivation behind our approach. We then describe each component of our system in detail.



Our first approach for addressing the contest specification was a novel end-to-end, deep learning-based system. The most challenging task was to find a way to represent the data; considering a sentence may have zero or many triples, and that the relations should be obtained directly from the text, it was exceedingly difficult to represent the input data in such a way that allowed the model to predict a decently-sized set of valid triples from a given document. Our best deep learning-based approach produced high-quality triples, but only in very small numbers. We hence decided to veer away from deep learning and capitalise on the wide variety of readily-available natural language processing tools.


For general English text, resources are available including several annotated benchmark datasets and off-the-shelf tools. For example,
CoNLL-2003 English benchmark dataset~\cite{sang2003conll2003} is a collection of Reuters news-wire articles, annotated with four entity types: persons, organizations, locations, and miscellaneous names\footnote{https://www.clips.uantwerpen.be/conll2003/ner/}. It contains around 300,000 tokens of 22,137 sentences.
OntoNotes5.0~\cite{weischedel2013ontonotes} is an annotated corpus of 2.9 million words from news, phone conversations, weblogs, broadcast, talk shows in three languages (English, Chinese, and Arabic) with structural information (syntax and predicate argument structure) and shallow semantics (word sense linked to an ontology and coreference)\footnote{https://catalog.ldc.upenn.edu/LDC2013T19}.
Off-the-shelf standard named entity recognition (NER) tools are able to recognize named entities of a restricted list of pre-defined entity types, such as location, person names, organization names, money, date, and time. Popular tools include NLTK~\cite{bird2009nltk}, SpaCy~\cite{honnibal2017spacy}, Stanford Named Entity Recogniser~\cite{finkel2005stanfordNER} and AllenNLP~\cite{Gardner2017AllenNLP,elmo2018peters}.

However, when it comes to real-world applications, such as the domain specific text in automotive engineering or public security, we face the low-resource data problem similar to machine translation between rare languages. There is no benchmark annotated dataset relevant to those domains, and it is near-impossible to find the right pivot language that allows us to take advantages of existing high resource NER tools. 
In automotive engineering domain, car types and car related names are more important than person or organisation names. For example, in the sentence \textit{Ford re-tuned the suspension and magnetic dampers to allow the GT350 to stiffen the suspension for better performance on the track}, the important entities are \textit{Ford}, \textit{GT350}, \textit{suspension}, and \textit{magnetic dampers}, but NER tools can only capture \textit{Ford} and \textit{GT350} as entities and ignore the other phrases. In order to avoid missing salient information units, chunking of noun phrases for entities and chunking of action related phrases for relations are performed in this work.

Our team also experimented with Open Information Extraction (OpenIE)~\cite{niklaus2018survey} and knowledge graph construction systems. There are a wide range of OpenIE systems available, with recent approaches incorporating neural networks in order to maximise performance~\cite{cui2018neural}. We found that OpenIE tends to produce a vast number of triples, with many subjects or objects being long sequences of words as opposed to useful entities. This is detrimental to the contest task, which demands a refined set of high-quality triples. Knowledge graph construction systems, such as T2KG~\cite{kertkeidkachorn2017t2kg}, rely on fixed relation types and as such are also undesirable for the contest task.


We ultimately found that the best performance was achieved by maintaining a high level of simplicity and utilising a pipeline-based approach. Our system is built using well-established natural language processing frameworks such as NLTK\footnote{https://www.nltk.org/} and SpaCy\footnote{https://spacy.io/}, and makes use of standard techniques such as tokenisation, part-of-speech (POS) tagging, named entity recognition, coreference resolution, and noun/verb phrase chunking. We incorporate several of our own algorithms in order to address the aforementioned shortcomings of NER on domain-specific data.

\subsection{Triple extraction system}

\label{subsec:triple-extraction-system}

\begin{figure}[!ht]
    \begin{center}
        \includegraphics[width=\linewidth]{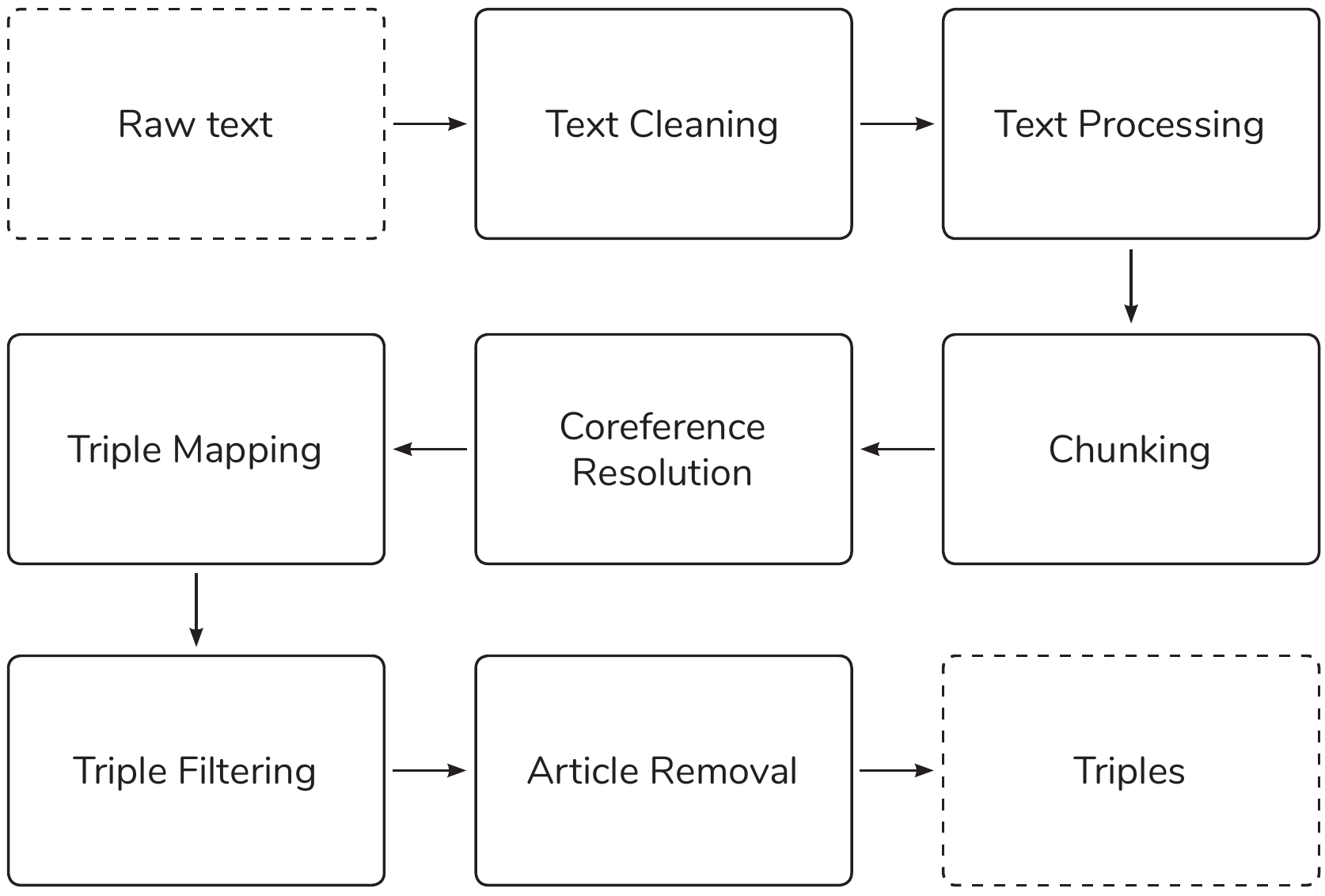}
        \caption{A diagram of the core components of our triple extraction system.}
        \label{fig:pipeline}
    \end{center}
\end{figure}

Our triple extraction system adopts a pipeline-based approach in order to convert a document into a set of triples. It comprises seven distinct stages, as shown in Figure~\ref{fig:pipeline}.

\textbf{Text cleaning}: Text data is cleaned to manage special characters such as hyphen and quotation marks and also break sentences joined together with no space between them.

\textbf{Text processing}: The text is processed through tokenisation, POS tagging, entity recognition and dependency parsing steps using SpaCy.
The results are shown in Table~\ref{tab:pos_tagging} for the following text:
\textit{Ford Motor Company is an American multinational automaker that has its main headquarters in Dearborn, Michigan, a suburb of Detroit. The company was founded by Henry Ford and incorporated on June 16, 1903.}

\begin{table*}[!h]
    \centering
    \begin{tabular}{clllllrrl}
    \hline
    Token Id & Token & Entity Type & IOB & Coarse Grained POS & POS & Start & End & Dependency \\
    \hline
    0 &  Ford &  ORG &  B &  PROPN &  NNP & 0& 3&  compound \\
    1&  Motor &  ORG &  I &  PROPN &  NNP & 5& 9&  compound \\
    2&  Company &  ORG &  I &  PROPN &  NNP & 11& 17&  nsubj \\
    3&  is &   &  O &  VERB &  VBZ & 19& 20&  ROOT \\
    4&  an &   &  O &  DET &  DT & 22& 23&  det \\
    5&  American &  NORP &  B &  ADJ &  JJ & 25& 32&  amod \\
    6&  multinational &   &  O &  ADJ &  JJ & 34& 46&  amod \\
    7&  automaker &   &  O &  NOUN &  NN & 48& 56&  attr \\
    8&  that &   &  O &  DET &  WDT & 58& 61&  nsubj \\
    9&  has &   &  O &  VERB &  VBZ & 63& 65&  relcl \\
    10&  its &   &  O &  DET &  PRP & 67& 69&  poss \\
    11&  main &   &  O &  ADJ &  JJ & 71& 74&  amod \\
    12&  headquarters &   &  O &  NOUN &  NN & 76& 87&  dobj \\
    13&  in &   &  O &  ADP &  IN & 89& 90&  prep \\
    14&  Dearborn &  GPE &  B &  PROPN &  NNP & 92& 99&  pobj \\
    15&  , &   &  O &  PUNCT &  , & 100& 100&  punct \\
    16&  Michigan &  GPE &  B &  PROPN &  NNP & 102& 109&  appos \\
    17&  , &   &  O &  PUNCT &  , & 110& 110&  punct \\
    18&  a &   &  O &  DET &  DT & 112& 112&  det \\
    19&  suburb &   &  O &  NOUN &  NN & 114& 119&  dobj \\
    20&  of &   &  O &  ADP &  IN & 121& 122&  prep \\
    21&  Detroit &  GPE &  B &  PROPN &  NNP & 124& 130&  pobj \\
    22&  . &   &  O &  PUNCT &  . & 131& 131&  punct \\
    23&  The &   &  O &  DET &  DT & 133& 135&  det \\
    24&  company &   &  O &  NOUN &  NN & 137& 143&  nsubjpass \\
    25&  was &   &  O &  VERB &  VBD & 145& 147&  auxpass \\
    26&  founded &   &  O &  VERB &  VBN & 149& 155&  ROOT \\
    27&  by &   &  O &  ADP &  IN & 157& 158&  agent \\
    28&  Henry &  PERSON &  B &  PROPN &  NNP & 160& 164&  compound \\
    29&  Ford &  PERSON &  I &  PROPN &  NNP & 166& 169&  pobj \\
    30&  and &   &  O &  CCONJ &  CC & 171& 173&  cc \\
    31&  incorporated &   &  O &  VERB &  VBD & 175& 186&  conj \\
    32&  on &   &  O &  ADP &  IN & 188& 189&  prep \\
    33&  June &  DATE &  B &  PROPN &  NNP & 191& 194&  pobj \\
    34&  16 &  DATE &  I &  NUM &  CD & 196& 197&  nummod \\
    35&  , &  DATE &  I &  PUNCT &  , & 198& 198&  punct \\
    36&  1903 &  DATE &  I &  NUM &  CD & 200& 203&  nummod \\
    37&  . &   &  O &  PUNCT &  . & 204& 204&  punct \\
    \hline
    \end{tabular}
    \caption{Text processing: tokenisation, POS tagging, entity recognition, and dependency parsing.}
    \label{tab:pos_tagging}
\end{table*}

\begin{table}[!ht]
    \centering
    \begin{tabular}{ccll}
    \hline
        Sent \# & Phrase \# & Phrase & Type \\
        \hline
        0 & 0 & Ford Motor Company & ENTITY \\
        0 &	1 &	is & VERB \\
        0 &	2 & an American multinational automaker & ENTITY \\
        0 &	3 &	that & DET \\
        0 &	4 &	has	& VERB \\
        0 &	5 &	its main headquarters & ENTITY \\
        0 &	6 &	in & ADP \\
        0 &	7 &	Dearborn & ENTITY \\
        0 &	8 &	, & PUNCT \\
        0 &	9 &	Michigan & ENTITY \\
        0 &	10 & , & PUNCT \\
        0 &	11 & a suburb of Detroit & ENTITY \\
        0 &	12 & . & PUNCT \\
        1 &	13 & The company & ENTITY \\
        1 &	14 & was founded by	& VERB \\
        1 &	15 & Henry Ford	& ENTITY \\
        1 &	16 & and	& CCONJ \\
        1 &	17 & incorporated on & VERB \\
        1 &	18 & June 16, 1903 & ENTITY \\
        1 &	19 & . & PUNCT \\
    \hline
    \end{tabular}
    \caption{Chunks of noun phrases and verb phrases.}
    \label{tab:chunks}
\end{table}

\textbf{Chunking}: Noun phrases (NPs) and verb phrases are chunked, as shown in Table~\ref{tab:chunks}. Noun chunks are phrases that have a noun and the words describing the noun. For example, \textit{an American multinational automaker} and \textit{a suburb of Detroit}.
We also implemented the chunking of action words, so that verb phrases can contain verbs, particles and/or adverbs that represent more meaningful relations between entities. For example, \textit{was founded by} and \textit{incorporated on}.

\begin{algorithm}
\caption{Chunking of noun phrases and verb phrases}\label{alg:chunking}
\begin{algorithmic}[1]
\Procedure{ChunkPhrases}{$document$}
\For{\textbf{each} $sentence$ $in$ $document$}
    \LeftComment{Chunk noun phrases (NPs) and tag as \textit{ENTITY}}
    \State \textbf{chunk} \textit{NPs} \Comment{NP}
        \State \textbf{chunk} $'(' + NP + ')'$ \Comment{(NP)}
        \State \textbf{chunk} $NP + 'of' + NP$ \Comment{NP of NP}
         \State \textbf{chunk} $NP + NP$ \Comment{NP NP}
    \LeftComment{Chunk verb phrases and tag as \textit{VERB}}
        \State \textbf{chunk} $VERB + PART$  \Comment{verb + particle}
        \State \textbf{chunk} $VERB + ADP$ \Comment{verb + adpositions} 
        \State \textbf{chunk} $ADP + VERB$ \Comment{adpositions + verb}
        \State \textbf{chunk} $PART + VERB$ \Comment{particle + verb}
        \State \textbf{chunk} $VERB + VERB$ \Comment{verb + verb} 
\EndFor
\State \textbf{return} $document$\Comment{Document with phrase chunks}
\EndProcedure
\end{algorithmic}
\end{algorithm}


\begin{algorithm*}
\caption{Triple mapping algorithm}\label{alg:triple}
\begin{algorithmic}[2]
\Procedure{GetTriples}{$document$}
\For{\textbf{each} $sentence$ $in$ $document$}
\State $relations \gets verbs + prepositions + postpositions$ \Comment{Select relations such as \textit{showcased, has, in, to, during}}
    \For{\textbf{each} $r$ $in$ $relations$}
    \State $heads\gets$ \textit{entities on the left side of $r$} \Comment{Get the head entities for the relation \textit{r}}
    \State $tails\gets$ \textit{entities on the right side of $r$} \Comment{Get the tail entities for the relation \textit{r}}
    \For{\textbf{each} $h$ $in$ $heads$}
    \For{\textbf{each} $t$ $in$ $tails$}
        \State $triples \gets triples + [h,r,t]$ \Comment{Add \textit{[head, relation, tail]} to the list of triples}
    \EndFor
    \EndFor
    \EndFor
\EndFor
\State \textbf{return} $triples$\Comment{Return the list of triples}
\EndProcedure
\Procedure{ExtractTriples}{$document$}
\LeftComment{Extract triples from the document at the sentence level}
\State $triples \gets \textsc{GetTriples}(document)$
\LeftComment{Extract the triples at the document level using the graph shortest paths}
\State $G \gets$ \textit{create graph(triples)} \Comment{Build a graph from the triples using NetworkX package}
\State $paths \gets$ \textit{get shortest paths(G)} \Comment{Get all shortest paths between named entities}
\For{\textbf{each} $h, t$ $in$ \textit{pairs of named entities}}
\If{\textit{h and t connected by a path using 'in', 'at', 'on' prepositions}}
    \State $triples \gets triples + [h,'in',t]$ \Comment{Add \textit{[head, 'in', tail]} to the list of triples}
\EndIf
\EndFor
\State \textbf{return} $triples$\Comment{Return the full list of triples}
\EndProcedure
\end{algorithmic}
\end{algorithm*}

\textbf{Coreference Resolution}: A list of coreferenced items is created using NeuralCoref\footnote{https://github.com/huggingface/neuralcoref}. For our example the following two coreference items are identified: \textit{Ford Motor Company} - \textit{its} and \textit{Ford Motor Company} - \textit{The company}. Coreference items are resolved on the triples by replacing the original phrase with the referred phrase for each item. For example, \textit{The company} will be replaced by \textit{Ford Motor Company}. In the case of pronouns such as \textit{its, her, his} or \textit{their}, we ignore the coreference items. As we prefer \textit{main headquarters} over \textit{Ford Motor Company main headquarters}, since \textit{main headquarters} will be connected to \textit{Ford Motor Company} by the triples.

\textbf{Triple Mapping}: Triples are created from the sentences in \textit{head, relation, tail} format using Algorithm~\ref{alg:triple}.
First, head and tail entities are extracted with their relations from the sentences and creates a list of triples.
Second, a graph is created from those triples to uncover the relations among named entities in separate sentences. Based on the relations of prepositions such as \textit{in, on, at}, more triples are created to provide more links between named entities in the graph.
Finally, the triples created by these two steps are joined to make the full list of triples for the given text.

\textbf{Triple Filtering}: To improve the quality of the triples, the filtering is performed to remove any triple with a stop word as a head entity. The stop words include NLTK stop words, names of days (Monday to Sunday) and names of months (January to December).

\textbf{Article Removal}: To clean the entities we removed some tokens including articles (e.g., a, an, the), possessive pronouns (e.g., its, their) and demonstrative pronouns (e.g., that, these) from the head and tails of each triple.

\subsection{Visualisation system}

\begin{figure}[!ht]
    \begin{center}
        \includegraphics[width=\linewidth]{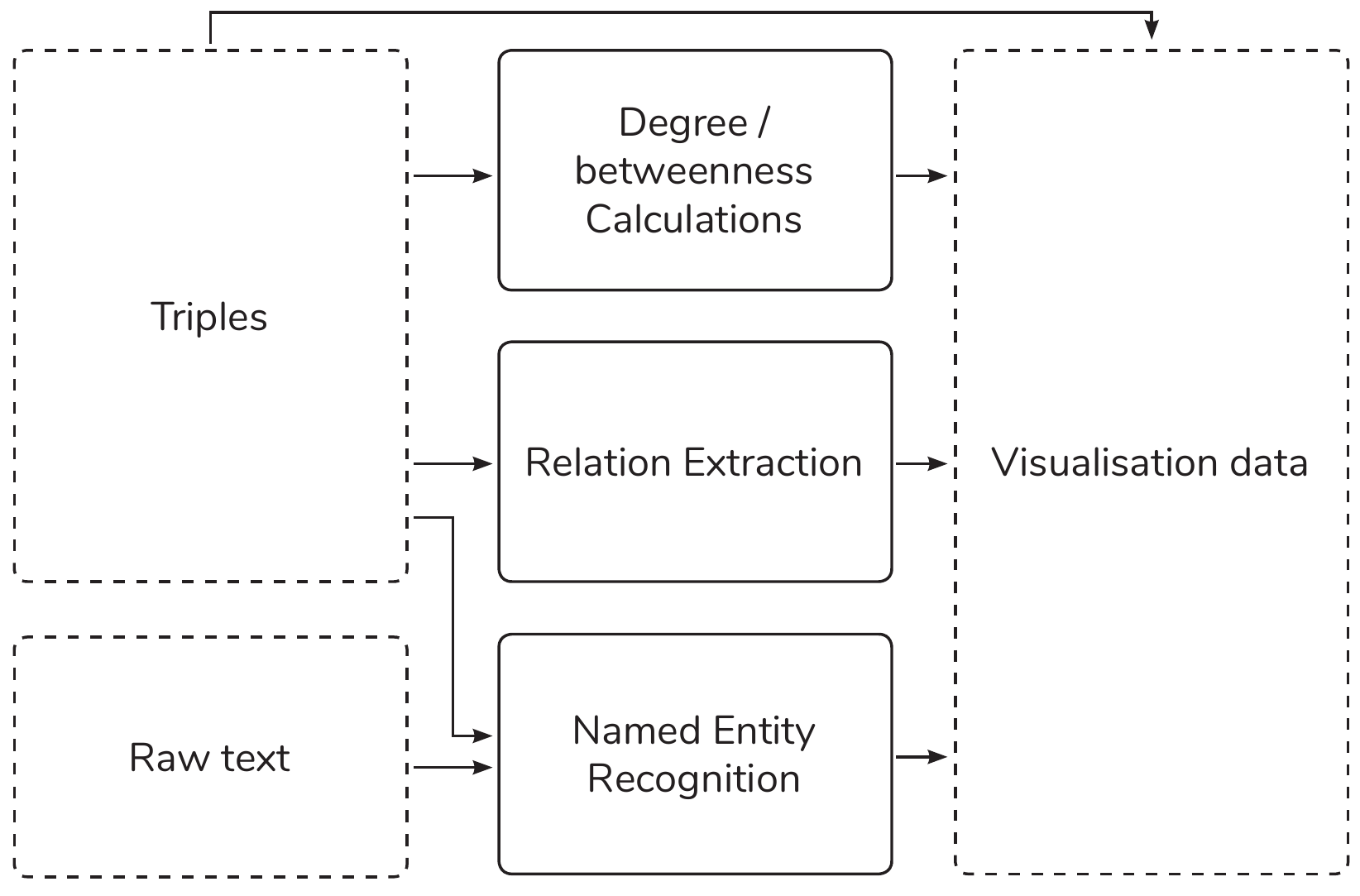}
        \caption{The additional stages performed by our system prior to visualisation in order to display more detailed information about each triple.}
        \label{fig:pipeline-vis}
    \end{center}
\end{figure}

Our visualisation system, which displays the results of the triple extraction system, performs three additional techniques in order to maximise the information displayed via our web application. After the triples have been generated as per Section~\ref{subsec:triple-extraction-system}, they are post-processed and appended with the degree/betweenness of the head and tail nodes, structured relation(s) corresponding to the verb relation of each triple, and the named entity classes of each head and tail. This process is displayed in Figure~\ref{fig:pipeline-vis}. The source code of our visualisation system is available on Github\footnote{https://github.com/Michael-Stewart-Webdev/text2kg-visualisation}.

The \textbf{degree/betweenness calculation} determines the degree and betweenness centrality of the head and tail of each triple. In graph theory, \textit{degree} refers to the number of edges connected to a node~\cite{diestel2005graph}. For triples, this directly corresponds to the number of triples in which each phrase appears. \textit{Betweenness centrality}, on the other hand, measures the extent to which each vertex lies along the paths between other vertices. Phrases that exert a high degree of influence over the flow of the graph, such as company names (``Ford'', ``BYD'') tend to have a high betweenness value and are hence more important than other terms. Incorporating the degree and betweenness calculations allows for this information to be conveyed in the visualisation.


The \textbf{relation extraction} component maps the relation phrase of each triple to one or more structured relation types. This allows for the graph visualisation to display structured relation types when desired by the user. Our system currently maps each relation phrase to its corresponding SemEval~\cite{hendrickx2009semeval} relation. To accomplish this we use an attention-based bidirectional Long Short-Term Memory (LSTM) model~\cite{zhou2016attention}, which maps a sequence of words padded with entity markers (\texttt{$\langle e_1 \rangle$} and \texttt{$\langle e_2 \rangle$}) to a fixed relation type. We create sequences using the head and tail of each triple as \texttt{$\langle e_1 \rangle$} and \texttt{$\langle e_2 \rangle$} respectively, and feed them into a pretrained model (trained on the SemEval 2010 Task 8 dataset) to obtain the corresponding SemEval relation. SemEval contains nine types of semantic relations and an additional type for other relations. 


Finally, the \textbf{named entity recognition} (NER) component determines the semantic type of the head and tail of each triple. We label each phrase with one of five types: \texttt{PER}, \texttt{ORG}, \texttt{LOC}, \texttt{MISC}, and \texttt{O}, based upon the Wikipedia NER  scheme~\cite{nothman2013learning}. The raw text is first labelled via SpaCy, yielding a set of entities $E$. Each phrase (head and tail) in each triple are then compared to every entity $e \in E$ and assigned the same label as $e$ when the phrase is highly similar to $e$ in terms of edit distance. One caveat of performing the NER after the triple extraction pipeline is that there is no contextual information passed to the named entity recognition model. However, applying NER immediately prior to visualisation allows for a greater level of abstraction and flexibility.

\section{Evaluation and Conclusion}

\begin{sidewaystable}[!ht]
{\small
    \centering
    \begin{tabular}{lll|lllllll}
    \hline
    \multicolumn{3}{c|}{Triple} &  \multicolumn{7}{c}{Additional information}  \\ \hline
    Head (\textsubscript{H})& Relation (\textsubscript{R}) & Tail (\textsubscript{T}) & SemEval Relation & Type\textsubscript{H} & Type\textsubscript{T} & Deg\textsubscript{H} & Deg\textsubscript{T} & Betw\textsubscript{H} & Betw\textsubscript{T}\\
    \hline
    Ford Motor Company &  in &  Dearborn & Content-Container &  ORG &  LOC & 6 & 3 & 11.0 & 0.75 \\
    Ford Motor Company &  in &  Michigan & Content-Container & ORG &  LOC & 6 & 3 & 11.0 & 0.75\\
    Ford Motor Company &  in &  suburb of Detroit & Member-Collection & ORG &  O & 6 & 3 & 11.0 & 0.75\\
    Ford Motor Company &  in &  June 16, 1903 &  Component-Whole & ORG &  O & 6 & 2 & 11.0 & 0.0\\
    Ford Motor Company &  is &  American multinational automaker & Instrument-Agency &   ORG &  O & 6 & 5 & 11.0 & 1.75\\
    Ford Motor Company &  was founded by &  Henry Ford & Product-Producer & ORG &  PER & 6 & 2 & 11.0 & 0.0\\
    American multinational automaker &  in &  Dearborn & Member-Collection & O &  LOC & 5 & 3 & 1.75 & 0.75\\
    American multinational automaker &  in &  Michigan & Member-Collection & O &  LOC & 5 & 3 & 1.75 & 0.75\\
    American multinational automaker &  in &  suburb of Detroit & Member-Collection &   O &  O & 5 & 3 & 1.75 & 0.75\\
    American multinational automaker &  has &  main headquarters & Cause-Effect &   O &  O & 5 & 4 & 1.75 & 1.0\\
    Henry Ford &  incorporated on &  June 16, 1903 & Component-Whole & PER &  O & 2 & 2 & 0.0 & 0.0\\
    main headquarters &  in &  Dearborn & Content-Container & O &  LOC & 4 & 3 & 1.0 & 0.75\\
    main headquarters &  in &  Michigan & Content-Container & O &  LOC & 4 & 3 & 1.0 & 0.75\\
    main headquarters &  in &  suburb of Detroit & Member-Collection & O &  O & 4 & 3 & 1.0 & 0.75\\
    
    \hline
    \end{tabular}
    \caption{Example triples produced by our triple extraction system, along with the additional information appended to each triple via our visualisation system.}
    \label{tab:triples}
}
\end{sidewaystable}

\subsection{Triple Extraction}

In order to evaluate the quality of our triple extraction system, we consider the following two sentences:
\textit{Ford Motor Company is an American multinational automaker that has its main headquarters in Dearborn, Michigan, a suburb of Detroit. It was founded by Henry Ford and incorporated on June 16, 1903.}

Table~\ref{tab:triples} displays the \textit{subject, predicate, object} triples from our triple extraction system and shows the additional information provided by the visualisation system:  the SemEval relation type, the named entity types of the heads and tails, and the degree and betweenness of each head and tail. 

The triples show some of the notable strengths of our model: the chunking component ensures useful phrases such as ``Ford Motor Company'' and ``Henry Ford'' appear in multiple triples. Furthermore, our system is able to extract useful triples with ``in'' relations via the triple mapping component, such as \texttt{(Ford Motor Company, in, Dearborn)}.

\subsection{Coreference Resolution}

\begin{figure}[!ht]
    \begin{center}
        \includegraphics[width=\linewidth]{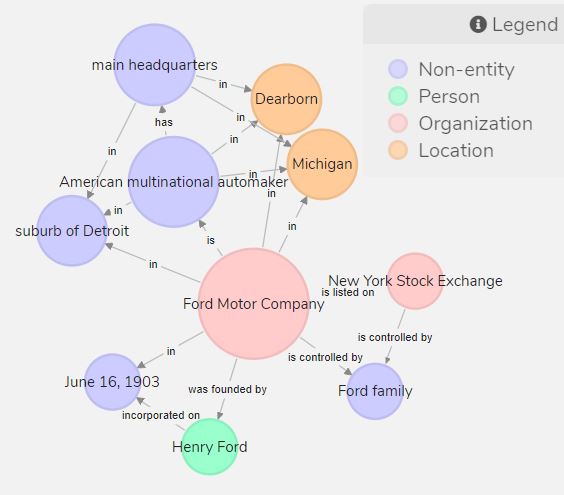}
        \caption{An example graph generated by our triple extraction system. The nodes are coloured based on their named entity types. The node sizes are based on their degree centrality values.}
        \label{fig:visualisation}
    \end{center}
\end{figure}

To highlight the effectiveness of our coreference resolution component, we introduce an additional sentence to our example so that it becomes: \\
\textit{\underline{Ford Motor Company} is an American multinational automaker that has its main headquarters in Dearborn, Michigan, a suburb of Detroit. \underline{It} was founded by Henry Ford and incorporated on June 16, 1903. \underline{The company} is listed on the New York Stock Exchange and \underline{it} is controlled by the Ford family.}

Figure~\ref{fig:visualisation} shows the result of visualising the above sentences via our web application. The underlined words (\underline{\textit{Ford Motor Company}} in sentence 1, \underline{\textit{It}} in sentence 2 and \underline{\textit{The company}} and \underline{\textit{it}} in sentence 3) represent the same entity \textit{Ford Motor Company}. The visualisation in Figure~\ref{fig:visualisation} clearly shows \textit{Ford Motor Company} as the shared entity between the three sentences. The node \textit{Ford Motor Company} appears the biggest among all nodes in the graph, to represent the highest degree centrality of the node in that graph.

\subsection{Conclusion}


In conclusion, our system uses a pipeline-based approach to extract a set of triples from a given document. It offers a simple and effective solution to the challenge of knowledge graph construction from domain-specific text. It also provides the facility to visualise useful information about each triple such as the degree, betweenness, structured relation type(s), and named entity types. 

It is important to note that the graph edit distance metric that is commonly used to automatically evaluate the quality of triples is only capable of structural analysis. In order to improve the metric it could be combined with meaningful semantic measures such as those present in the machine translation and image captioning domains (e.g. SPICE~\cite{anderson2016spice}).  Another option would be to incorporate a simple sum of word embeddings over each triple so that semantic information is captured by the metric.

In future we plan to continue working on our end-to-end deep learning-based triple extraction model. 
\section{External Resources}

Our triple extraction system uses the aforementioned NLTK~\cite{bird2009nltk} and SpaCy~\cite{honnibal2017spacy} at various stages throughout the pipeline. 

Our visualisation system is written in Flask\footnote{https://www.fullstackpython.com/flask.html}. The front-end visualisations are written primarily in D3.js\footnote{https://d3js.org/}. The attention-based Bi-LSTM~\cite{zhou2016attention} for relation extraction is implemented in Tensorflow~\cite{abadi2016tensorflow}, and trained on the SemEval 2010 Task 8 dataset~\cite{hendrickx2009semeval}. The degree and betweenness calculations are performed via NetworkX\footnote{https://networkx.github.io}.


\section*{Acknowledgement}

We would like to thank our team members Morgan Lewis and Thomas Smoker, who are in the early stage of their PhD candidatures, for their contributions on literature search.
\clearpage
\bibliographystyle{./bibliography/IEEEtran}
\bibliography{bibliography.bib}

\end{document}